\documentclass[conference]{IEEEtran}
\IEEEoverridecommandlockouts
\usepackage{cite}
\usepackage{amsmath,amssymb,amsfonts}
\usepackage{algorithmic}
\usepackage{algorithm} 
\usepackage{graphicx}
\usepackage{textcomp}
\usepackage{xcolor}
\usepackage{tabularx}

\usepackage{array}  

\newcolumntype{Y}{>{\raggedright\arraybackslash}X}
\def\BibTeX{{\rm B\kern-.05em{\sc i\kern-.025em b}\kern-.08em
    T\kern-.1667em\lower.7ex\hbox{E}\kern-.125emX}}
\begin{document}

\title{
S2S-FDD: Bridging Industrial Time Series\\ and Natural Language for Explainable\\ Zero-shot Fault Diagnosis
\thanks{This work is supported by the Zhejiang Key Research and Development Project (2024C01163), the National Natural Science Foundation of China (No. 62125306), the National Natural Science Foundation of China (No. 62450020), the State Key Laboratory of Industrial Control Technology, China (ICT2025C01), and the Open Research Project of the State Key Laboratory of Industrial Control Technology, China (ICT2025B27).}
}

\author{\IEEEauthorblockN{1\textsuperscript{st} Baoxue Li}
\IEEEauthorblockA{\textit{1. State Key Laboratory of Industrial Control Technology,} \\ \textit{College of Control Science and Engineering} \\
\textit{Zhejiang University}\\
Hangzhou, China}
\and
\IEEEauthorblockN{2\textsuperscript{nd} Chunhui Zhao*}
\IEEEauthorblockA{\textit{1. State Key Laboratory of Industrial Control Technology,} \\ \textit{College of Control Science and Engineering} \\
\textit{Zhejiang University}\\
\textit{2. School of Information and Electrical Engineering,} \\
\textit{Zhejiang University City College} \\
Hangzhou, China \\
chhzhao@zju.edu.cn}
}

\maketitle

\begin{abstract}
Fault diagnosis is critical for the safe operation of industrial systems. Conventional diagnosis models typically produce abstract outputs such as anomaly scores or fault categories, failing to answer critical operational questions like ``Why" or ``How to repair". While large language models (LLMs) offer strong generalization and reasoning abilities, their training on discrete textual corpora creates a semantic gap when processing high-dimensional, temporal industrial signals. 
To address this challenge, we propose a Signals-to-Semantics fault diagnosis (S2S-FDD) framework that bridges high-dimensional sensor signals with natural language semantics through two key innovations:
We first design a Signal-to-Semantic operator to convert abstract time-series signals into natural language summaries, capturing trends, periodicity, and deviations.
Based on the descriptions, we design a multi-turn tree-structured diagnosis method to perform fault diagnosis by referencing historical maintenance documents and dynamically querying additional signals. The framework further supports human-in-the-loop feedback for continuous refinement. 
Experiments on the multiphase flow process show the feasibility and effectiveness of the proposed method for explainable zero-shot fault diagnosis.
\end{abstract}


\section{Introduction}
Fault diagnosis is essential to ensuring the safe and stable operation of industrial processes \cite{fault}. With the advancement of sensing technologies in modern industry, massive operational data can now be collected, paving the way for data-driven approaches \cite{conditiondriven}. These methods can identify abnormal patterns directly from historical data, enabling the monitoring and diagnosis of complex systems that are difficult to model analytically.

Over the past decades, both multivariate statistical methods and deep learning techniques have seen success in this field. Statistical methods such as Fisher Discriminant Analysis (FDA) \cite{fda}, Support Vector Machines (SVM) \cite{svm}, and Bayesian Networks \cite{bayes} have laid a solid foundation. More recently, deep learning models, particularly Convolutional Neural Networks (CNNs) \cite{cnn} and Transformers \cite{transformer}, have pushed the boundaries of fault detection \cite{survey_monitor2}. Given the scarcity of fault samples in industrial environments, researchers have explored few-shot learning \cite{few} and transfer learning \cite{transfer} to improve diagnosis performance. In parallel, the demand for interpretability in industrial settings has led to the development of explainable diagnosis algorithms \cite{xai}. Despite these advancements, most existing approaches still require fault data during training. Feng \cite{zsl} introduced the concept of zero-shot fault diagnosis and provided a theoretical framework for its feasibility. This method leverages attribute transfer to recognize unseen faults while improving interpretability through the generation of attribute vectors during prediction.
Therefore, zero-shot fault diagnosis has recently emerged as a promising direction \cite{zsl2}. However, conventional diagnosis models typically produce abstract outputs such as anomaly scores or fault categories without answering critical questions like “Why is this abnormal?” or “How should we repair it?”

The rise of large language models (LLMs) has brought transformative progress across various domains \cite{llm_survey}. These models exhibit strong comprehension and generalization capabilities, offering a potential path toward zero-shot and interpretable fault diagnosis \cite{llm_fault,zsl3}. 
However, they have not yet achieved the same level of success in industrial applications. 
One major obstacle lies in the nature of industrial data—high-dimensional, abstract, and temporal—which poses challenges to current large models primarily trained on discrete, token-based textual corpora. 
Unlike natural language, time-series data are continuous and dynamic, making them difficult to discretize or encode in ways that language models can effectively interpret \cite{timellm}. 
In addition to temporal understanding, conventional models often lack domain-specific knowledge of industrial processes \cite{know}, raising concerns about the reliability and stability of their outputs in real-world diagnosis tasks.

To address these challenges, we propose a Signals-to-Semantics (S2S) framework for industrial fault diagnosis that can operate without any fault data. 
Our approach is capable of interpreting abstract time-series signals and producing reliable diagnosis results by leveraging historical maintenance records. 
Central to our design is a reconstruction-based S2S operator that enables the model to assess deviations between current and baseline (normal) signals. 
This operator transforms numerical time-series inputs into natural language descriptions with trends, periodicity, and deviations in industrial terms.

Building on these textual representations, we design a multi-turn tree-structured diagnosis method powered by LLMs. The method can retrieve relevant historical maintenance documents, perform iterative reasoning, and dynamically request additional sensor measurements to address input information gaps. More importantly, the framework supports human-in-the-loop feedback, allowing experts to refine the reasoning process and establish a closed-loop optimization system.

The contributions of our work are summarized as follows.

\begin{enumerate}
\item{
We identify a fundamental challenge in intelligent industrial maintenance: the
semantic gap between industrial time-series data and natural language understanding. To address this, we propose a Signals to Semantics (S2S) framework, translating sensor data into semantically rich descriptions by leveraging LLMs.
}

\item{
To bridge the gap between continuous time-series signals and natural language understanding, we design a Signals-to-Semantics operator that converts raw sensor data into domain-aware natural language summaries, capturing trends, periodicity, and deviations.
}

\item{
We propose a multi-turn tree-structured diagnosis method based on LLMs. Leveraging textual time-series descriptions, the method achieves zero-shot fault diagnosis and analysis. It also supports dynamic reasoning updates based on operator feedback, forming a human-in-the-loop adaptive diagnosis loop.
}

\end{enumerate}

\section{Method}

In this section, the details of the proposed S2S framework are presented. It contains two key components, including the S2S operator and multi-turn tree-structured diagnosis method. The former converts raw sensor data into concise and domain-aware natural language summaries, bridging the gap between continuous time-series signals and natural language understanding. Based on the descriptions, the latter retrieves relevant historical maintenance documents and conducts zero-shot fault diagnosis.

\begin{figure*}[htbp]
\centering
\includegraphics[width=0.8\linewidth]{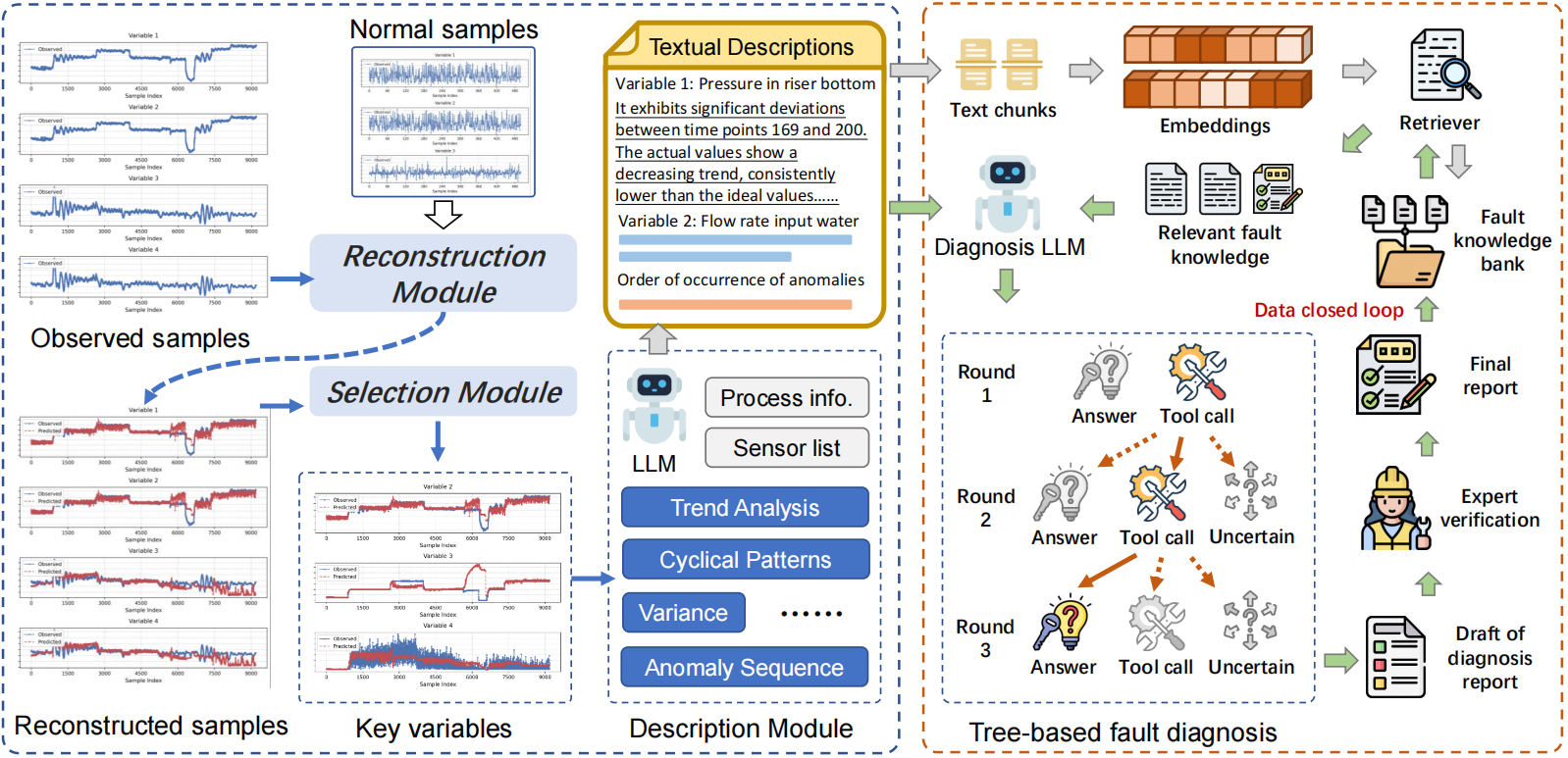}
\caption{Overview of the proposed S2S framework. It contains two key components, including the S2S operator and multi-turn tree-structured diagnosis method. The former converts raw sensor data
into concise and domain-aware natural language summaries. Based on the descriptions, the latter retrieves relevant historical maintenance documents and conducts zero-shot fault diagnosis.}
\label{fig:framework}
\end{figure*}

\subsection{The Signals to Semantics operator} 

In order to better obtain an accurate tim description, we introduce an S2S operator. This operator is essentially a reconstruction module that satisfies the logic of the specialist to determine the faults, i.e., to compare them with the normal values. This comparison helps the LLM to output a more reliable and targeted description.

For an industrial system with \( m \) measurement points (e.g., temperature, pressure), the sample collected at time \( i \) is denoted as \( \mathbf{W}_i = [x_1(i), x_2(i), \dots, x_m(i)]^T \in \mathbb{R}^{m \times 1} \). Temporal data under normal operating conditions are collected to construct a feature matrix \( \mathbf{W} \in \mathbb{R}^{m \times L} \), where \( L \) is the length of the time series. To characterize typical temporal patterns, \( n \) representative samples are selected via the following steps:  
(1). Cluster the \( L \) time-series samples into \( n \) clusters using K-means clustering.  
(2). Calculate the geometric centroid for each cluster.  
(3). Select the sample closest to each centroid as a representative sample, which is then aggregated into a state matrix:  
\begin{equation*}
\begin{aligned}
\mathbf{D} &= [\mathbf{W}_1, \mathbf{W}_2, \dots, \mathbf{W}_n] \\
&= 
\begin{bmatrix}
x_1(1) & x_1(2) & \dots & x_1(n) \\
x_2(1) & x_2(2) & \dots & x_2(n) \\
\vdots & \vdots & \ddots & \vdots \\
x_m(1) & x_m(2) & \dots & x_m(n) \\
\end{bmatrix} 
\in \mathbb{R}^{m \times n}  
\end{aligned}
\end{equation*}

This matrix represents \( n \) typical temporal patterns of the system under normal operation.  

For an online input sample \( \mathbf{W}_{\text{in}} \in \mathbb{R}^{m \times 1} \), a weight vector \( \boldsymbol{\omega} \in \mathbb{R}^{n \times 1} \) is computed to represent \( \mathbf{W}_{\text{in}} \) as a linear combination of the normal temporal patterns in \( \mathbf{D} \). The weight vector is derived via:  
\begin{equation}
\boldsymbol{\omega} = (\mathbf{D}^T \mathbf{D})^{-1} \mathbf{D}^T \mathbf{W}_{\text{in}}
\end{equation}  
where \( (\cdot)^{-1} \) denotes matrix inversion. 
This formula can be derived from least squares regression directly.
The input sample is reconstructed as \( \mathbf{W}_{\text{out}} = \mathbf{D} \boldsymbol{\omega} \). A large reconstruction residual \( \text{RES} = \mathbf{W}_{\text{in}} - \mathbf{W}_{\text{out}} \) indicates a potential fault.

Given test data \( \mathbf{X} \) and its reconstruction residual \( \mathbf{R} \), both are segmented into baseline and fault portions using the fault start time \( t_{\text{start}} \) and end time \( t_{\text{end}} \) (provided by the monitoring system with allowable deviations):  
\begin{equation}
\begin{aligned}
\mathbf{X}_{\text{base}} &= \mathbf{X}[: t_{\text{start}}] \in \mathbb{R}^{t_{\text{start}} \times m} \\
\mathbf{X}_{\text{fault}} &= \mathbf{X}[t_{\text{start}}: t_{\text{end}}] \in \mathbb{R}^{(t_{\text{end}} - t_{\text{start}} + 1) \times m} \\
\mathbf{R}_{\text{base}} &= \mathbf{R}[: t_{\text{start}}] \in \mathbb{R}^{t_{\text{start}} \times m} \\
\mathbf{R}_{\text{fault}} &= \mathbf{R}[t_{\text{start}}: t_{\text{end}}] \in \mathbb{R}^{(t_{\text{end}} - t_{\text{start}} + 1) \times m}
\end{aligned}
\end{equation}  

The average baseline reconstruction error is:  
\begin{equation}
b = \frac{1}{t_{\text{start}}} \sum_{i=1}^{t_{\text{start}}} |r_i|
\end{equation}  
For each variable \( j \), the absolute residual in the fault segment is \( d_j = |\mathbf{R}_{\text{fault}}[:, j]| \), and the anomaly threshold is \( \tau_j = \alpha \cdot b_j \), where \( \alpha \) is a pre-set coefficient and \( b_j \) is the baseline residual for variable \( j \).  

An anomaly indicator function identifies faulty time points:  
\begin{equation}
d_j(t) = \begin{cases}
1, & \text{if } d_j(t) \geq \tau_j \\
0, & \text{otherwise}
\end{cases}
\end{equation}  

A fault is confirmed if the indicator remains 1 for a consecutive window of length \( W \). The earliest fault occurrence time for variable \( j \) is:  
\begin{equation}
t_j^* = \min \left\{ t \mid \sum_{i=t}^{t+W-1} d_j(i) = W \right\}
\end{equation}  
The anomaly score for variable \( j \) is:  
\begin{equation}
s_j = \begin{cases}
0, \text{if } \{d_j(t) \mid d_j(t) \geq \tau_j\} = \emptyset\\
\left( \dfrac{\operatorname{mean}(\{d_j(t) \mid d_j(t) \geq \tau_j\})}{b_j} - 1 \right) \times 100, \text{otherwise}\\

\end{cases}
\end{equation}

Candidate variables are selected by top \( n_1 \) anomaly scores (\(\mathcal{S}_1\)) and top \( n_2 \) earliest fault times (\(\mathcal{S}_2\)), forming \( \mathcal{S} = \mathcal{S}_1 \cup \mathcal{S}_2 \). Final selection uses variance comparison:  
\begin{equation}
\mathcal{S}_{\text{filtered}} = \{ j \in \mathcal{S} \mid \text{var}(\mathbf{X}_{\text{fault}}[:, j]) > 2 \cdot \text{var}(\mathbf{X}_{\text{base}}[:, j]) \}
\end{equation}  

For any target variable \( j \in S_{\text{filtered}} \), there exists a corresponding data table \( \text{Table}_j \).
The measured values, reconstructed values, reconstruction errors, and reconstruction error percentages of the target variables obtained in \(\mathcal{S}_{\text{filtered}}\) with length $t_{end}-t_{start}$  are organized into tabular form.  The final constructed temporal description prompt can be seen in Table \ref{tab:temporal_prompt}.

\begin{table*}[htbp]
\centering
\caption{Prompt Templates}
\scriptsize
\begin{tabularx}{\textwidth}{|X|}
\hline
\textbf{Prompt Template for Temporal Description} \\
\hline
 
Your task is to describe the deviation between the measured value and the ideal normal value of a measurement point, focusing on trends, periodic patterns, volatility, and key anomalies.
\begin{itemize}
  \item Industrial process: \texttt{[PROCESS\_INFO]}
  \item All measurement points: \texttt{[ALL\_SENSORS]}
  \item Target measurement point: \texttt{[TARGET\_SENSOR]}
  \item Data table: \texttt{[TABLE]} (measured value, ideal normal value, deviation, and time-varying deviation percentage, along with the average deviation and deviation percentage under normal conditions)
\end{itemize} 
Only focus on time intervals where the deviation or deviation percentage significantly exceeds that under normal conditions, as these intervals indicate possible faults. Ignore periods with deviation close to normal. For significantly deviated periods, describe:
\begin{enumerate}
  \item The trend of the measured value (increasing, decreasing, stable, or cyclically fluctuating)
  \item Whether the measured value is above or below the ideal value
\end{enumerate}
If no significant deviation is observed, state that the variable has no obvious abnormality. Provide both quantitative indicators (time intervals) and qualitative insights, focusing on observable patterns. Avoid speculating on root causes. Do not use subheadings or Markdown syntax. Keep the response concise—no more than 100 words. \\
\hline
\textbf{Prompt Template for Fault Diagnosis}\\
\hline
You are an expert in fault diagnosis for industrial processes. Your task is to analyze the following information to identify potential faults:
\begin{itemize}
  \item Industrial Process: \texttt{\detokenize{[PROCESS_INFO]}}
  \item Measurement Points (List of available sensors): \texttt{\detokenize{[ALL_SENSORS]}}
  \item Fault Knowledge (Known faults and their characteristics): \texttt{\detokenize{[FAULT_KNOWLEDGE]}}
  \item Time-Series Observations: \texttt{\detokenize{[TIME_DESP]}} (Deviation = Measured - Predicted)
\end{itemize}

Follow these steps:
\begin{enumerate}
  \item Identify Key Sensors: Determine which sensors show significant deviations from expected values.
  \item Cross-Reference with Fault Knowledge: Compare the observed patterns with known fault characteristics to hypothesize possible faults. Please note that sometimes different faults may exhibit similar patterns in certain variables. So, do not rely solely on the provided fault knowledge; you should combine fault causes and process workflow to reasonably infer the trends of different variables.
  \item Check data sufficiency:
    \begin{itemize}
      \item If the time series observations provide enough information to identify the fault, output the fault number in the following format:\texttt{<reasoning>...</reasoning>} and \texttt{<answer>...</answer>}
      \item If the time series observations are insufficient or critical variables are missing, use the `get\_target\_table` tool to query detailed data for specific sensors. Ensure the sensor name exists in the Measurement Points before querying. Output the tool call in the following format:\texttt{<tool>get\_target\_table("SENSOR")</tool>}
    \end{itemize}
\end{enumerate}

Important Notes:
\begin{itemize}
  \item Prioritize sensors listed in \texttt{\detokenize{[ALL_SENSORS]}}
  \item Only call \texttt{get\_target\_table} when necessary:
    \begin{itemize}
      \item When time series observations are insufficient to determine the fault
      \item When critical variables for fault diagnosis are missing from the observations
    \end{itemize}
\end{itemize}\\
\hline
\end{tabularx}\label{tab:temporal_prompt}
\end{table*}

In the prompt, \texttt{[PROCESS\_INFO]} denotes the textual description of the background and principles of the industrial process, \texttt{[ALL\_SENSORS]} denotes all the measurement points of the industrial object, i.e., the variable information, \texttt{[TARGET\_SENSOR]} denotes the target measurement point, i.e., the target variable in $S_{\text{filtered}}$, and \texttt{[TABLE]} denotes the target variable corresponding to the data table $\text{Table}_j$.

The integrated prompt is input into the large language model, which will sequentially generate temporal textual descriptions \( \mathcal{D}_i \) for all target variables (\( i = 1, 2, 3, \ldots, p \)), where \( p = |S_{\text{filtered}}| \) denotes the total number of target variables.

\subsection{Multi-turn tree-structured diagnosis method}
In this part, the details of multi-turn tree-structured diagnosis method are introduced.
This part includes two important aspects: how to retrieve relevant knowledge from the historical knowledge base, and how to combine the retrieved knowledge for diagnosis.

First, fault knowledge base from diagnosis records and expert experience is encoded into embeddings \( \mathbf{e}_j = \text{Embedder}(K_j) \), where \( K_j \) are fault records. LLM-generated descriptions \( D_i \) are also embedded (\(\mathbf{d}_i\)), and their cosine similarity with \( \mathbf{e}_j \) is computed:  
\begin{equation}
\text{sim}(D_i, K_j) = \frac{\mathbf{d}_i \cdot \mathbf{e}_j}{\|\mathbf{d}_i\| \|\mathbf{e}_j\|}
\end{equation}  
Fault records $K_{target}$ with similarity above a threshold are included in LLM prompts to activate reasoning. Notably, a chunking operation is performed before encoding each report \( K_j \), and after recalling a particular chunk, the corresponding full report is retrieved.

Then, the fault diagnosis prompt is constructed based on textual descriptions of industrial processes, sensor measurement point information, relevant fault descriptions $K_{target}$, and temporal textual descriptions of target variables \( \mathcal{D}_i \), and the prompt is fed into the LLM for the fault diagnosis. 
The prompt can be seen in Table \ref{tab:temporal_prompt}.
If the LLM feels that the textual information for given variables is not sufficient to infer a fault, a function calling is made to retrieve information about variables that the LLM believes may be useful for fault diagnosis. 
Therefore, the diagnosis process is a tree-based structure as shown in the right part of Fig. \ref{fig:framework}.
Notably, the results of the funtion calling will be appended into the massages and the model will be prompted to continue the diagnosis, allowing it to output the \texttt{<uncertain>} mode in addition to the \texttt{<answer>} and \texttt{<tool>} modes.
The output of the LLM is reviewed by several votes, and the fault type with the most votes is selected as the final result, while the inference process of the LLM output is used as the judgment basis.
The whole process can be found in Algorithm \ref{alg1}.

Moreover, the results of the model diagnosis can be further organized to output a fault diagnosis report, which can be incorporated into the knowledge base for future retrieval after dedicated expert verification. This constitutes a human-in-the-loop closure and knowledge iteration.

\subsection{Theoretical Analysis of Fault Detectability}
The proposed reconstruction module leverages the linear representation of normal temporal patterns to distinguish between healthy and faulty operations. Below, we provide a theoretical theorem to analyze the fault detectability.

\textbf{Theorem 1 (Residual Energy Bound and Detectability).}

Let the state matrix \(\mathbf{D} \in \mathbb{R}^{m \times n}\) span a subspace \(\mathfrak{M} \subset \mathbb{R}^m\) with \(\sigma_{\min}(\mathbf{D})\) denoting its minimum singular value. For a faulty sample \(\mathbf{W}_{\text{fault}} = \mathbf{D}\boldsymbol{\omega} + \boldsymbol{\delta}\), the residual energy satisfies:  
\[
\|\text{RES}\|_2^2 = \|\boldsymbol{\delta}_{\perp}\|_2^2 = \|\boldsymbol{\delta}\|_2^2 \sin^2\theta\geq \frac{\sigma_{\min}(\mathbf{D})\|\boldsymbol{\delta}\|_2}{\sqrt{\sigma_{\min}^2(\mathbf{D}) + \sigma_{\max}^2(\mathbf{D})}} ,
\]  
where \(\boldsymbol{\delta}_{\perp}\) is the orthogonal projection of \(\boldsymbol{\delta}\) onto \(\mathfrak{M}^\perp\), \(\theta\) is the principal angle between \(\boldsymbol{\delta}\) and \(\mathfrak{M}\), and \(\sigma_{\max}(\mathbf{D})\) is the maximum singular value of \(\mathbf{D}\).

\textbf{Proof.}
The residual is the orthogonal projection of \(\boldsymbol{\delta}\):
\[
\text{RES} = \mathcal{P}_{\mathfrak{M}^\perp}(\mathbf{W}_{\text{fault}} - \mathbf{D}\boldsymbol{\omega}) = \mathcal{P}_{\mathfrak{M}^\perp}(\boldsymbol{\delta}) = \boldsymbol{\delta}_{\perp}.
\]
  
By the definition of principal angles in Euclidean space:
\[
\sin\theta = \frac{\|\boldsymbol{\delta}_{\perp}\|_2}{\|\boldsymbol{\delta}\|_2} \implies \|\text{RES}\|_2 = \|\boldsymbol{\delta}\|_2 \sin\theta.
\]
 
According to standard results in subspace perturbation theory, the sine of the principal angle between a vector $\boldsymbol{\delta}$ and the subspace $\mathfrak{M}$ satisfies:

$$
\sin\theta \geq \frac{\sigma_{\min}(\mathbf{D})}{\sqrt{\sigma_{\min}^2(\mathbf{D}) + \sigma_{\max}^2(\mathbf{D})}}.
$$

Therefore, we obtain the residual energy lower bound:

$$
\|\text{RES}\|_2 = \|\boldsymbol{\delta}\|_2 \sin\theta \geq \|\boldsymbol{\delta}\|_2 \cdot \frac{\sigma_{\min}(\mathbf{D})}{\sqrt{\sigma_{\min}^2(\mathbf{D}) + \sigma_{\max}^2(\mathbf{D})}}.
$$

This completes the proof. $\blacksquare$

\begin{algorithm}
\caption{The S2S Framework For Fault Diagnosis}
\begin{algorithmic}[1]
\label{alg1}
\footnotesize
\STATE \textbf{Initialization}:
\STATE Set the maximum retry count as $R_{\text{max}}$.
\STATE Set the retry counter as $r \gets 0$.
\STATE Set the message list as $\mathcal{M} \gets \{\text{User Input}\}$.

\STATE \textbf{Main Loop}:
\WHILE{$r < R_{\text{max}}$}
    \STATE \textbf{Chat with LLM}:
    \STATE Obtain model response $\mathbf{R} \gets \text{Model}(\mathcal{M})$.
    \STATE Extract response content $\mathbf{R}_{\text{content}}$, update message list: $\mathcal{M} \gets \mathcal{M} \cup \{\mathbf{R}\}$.

    \STATE \textbf{Check Response Pattern}:
    \IF{$\mathbf{R}_{\text{content}}$ contains \texttt{<answer>}}
        \STATE Extract fault $f \gets \text{Extract}(\mathbf{R}_{\text{content}}, \texttt{<answer>})$.
        \STATE Store result: $\text{Result} \gets f$, exit loop.
    \ELSIF{$\mathbf{R}_{\text{content}}$ contains \texttt{<tool>}}
        \STATE Parse function calling: $\mathcal{T} \gets \text{Parse}(\mathbf{R}_{\text{content}}, \texttt{<tool>})$.
        \STATE Validate function calling legality: $\mathcal{T}_{\text{valid}} \gets \{t \in \mathcal{T} \mid \text{Validate}(t)\}$.
        \STATE Execute function calling and obtain results: $\mathcal{R}_{\text{tool}} \gets \{\text{Execute}(t) \mid t \in \mathcal{T}_{\text{valid}}\}$.
        \STATE Construct tool result prompt: $\mathbf{P} \gets \text{ConstructPrompt}(\mathcal{R}_{\text{tool}})$.
        \STATE Update message list: $\mathcal{M} \gets \mathcal{M} \cup \{\mathbf{P}\}$, continue loop.
    \ELSIF{$\mathbf{R}_{\text{content}}$ contains \texttt{<uncertain>}}
        \STATE Extract possible fault number list: $\mathcal{F} \gets \text{Extract}(\mathbf{R}_{\text{content}}, \texttt{<uncertain>})$.
        \STATE Store result: $\text{Result} \gets \mathcal{F}$, exit loop.
    \ELSE
        \STATE $r \gets r + 1$.
    \ENDIF
\ENDWHILE

\STATE \textbf{Termination Condition}:
\IF{$r = R_{\text{max}}$}
    \STATE Output ``Maximum retry count reached" and set default result: $\text{Result} \gets 0$.
\ENDIF
\end{algorithmic}
\end{algorithm}

\section{Experiments}

\begin{figure}[htbp]
\centering
\includegraphics[width=0.8\linewidth]{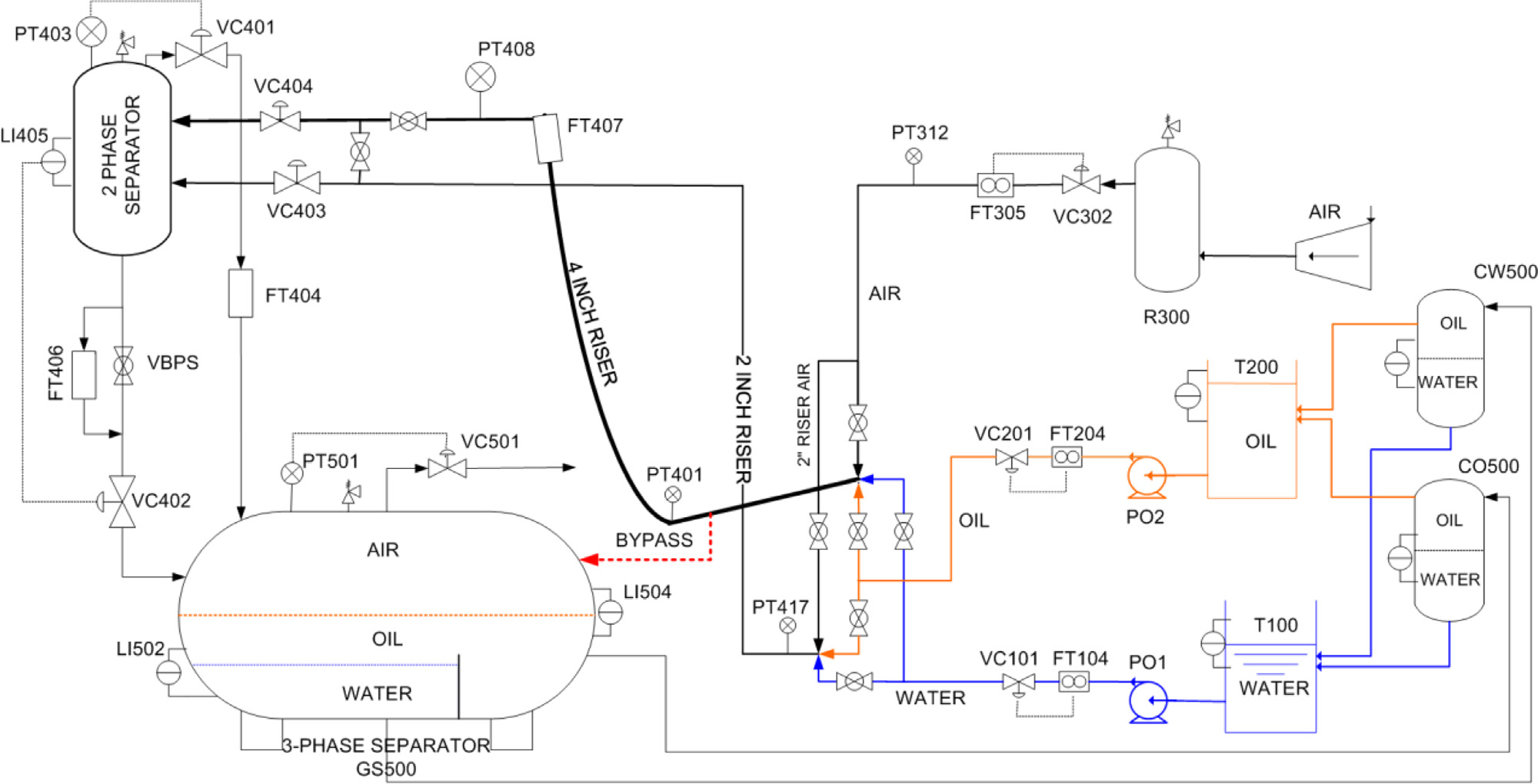}
\caption{Sketch of the multiphase flow process.}
\label{fig:phase}
\end{figure}

In this section, the performance of the proposed TSA-ILLM is evaluated on the multiphase flow process.
The multiphase flow facility at Cranfield University is designed to provide a controlled and measured flow rate of water, oil, and air to a pressurized system \cite{phase}. It can be supplied with a single phase of air, water, and oil, or a mixture of those fluids at certain rates.
The sketch of the multiphase flow process is shown in Fig. \ref{fig:phase}.

\subsection{Experimental setup}
We conducted experiments on both LLMs and reasoning LLMs. Qwen2.5-7B, 72B, Deepseek-V3 were selected in the non-reasoning LLMs.
DeepSeek-R1-Distill-Qwen-7B, DeepSeek-R1-Distill-Qwen-32B, QwQ-32B, DeepSeek-R1 were selected in the reasoning LLMs.

The multiphase flow process includes six fault types.
In this paper, five fault types are chosen and the fault 4 is excluded.
Fault 4 is simulating a leak at the bottom of the riser by opening direct bypass. Since the bottom of the riser is not really leaking, there is no reduction in flow through the entire system, and some of the variable curves do not behave exactly like the leak. Therefore, we removed this fault.
500 normal samples are used to construct the state matrix.
For information that needs to be filled in the prompts, such as \texttt{\detokenize{[PROCESS_INFO]}}, \texttt{\detokenize{[ALL_SENSORS]}}, and \texttt{\detokenize{[FAULT_KNOWLEDGE]}} are extracted from the paper \cite{phase}.

A total of 13 cases were evaluated with fault 1 (case 1-case 3), fault 2 (case 4-case 6), fault 3 (case 7-case 9), fault 5 (case 10-case 11), and fault 6 (case 12-case 13).
Each case was reasoned 5 times and the final diagnosis result was voted to obtain and calculate the accuracy.

\subsection{Comparison of diagnosis accuracy}

\begin{table*}[h]
\centering
\setlength{\tabcolsep}{2pt}
\caption{Comparison of Different Methods}
\begin{tabular}{l|ccc|ccc|ccc|cc|cc|c}
\hline
\textbf{Methods} & \textbf{Case 1} & \textbf{Case 2} & \textbf{Case 3} & \textbf{Case 4} & \textbf{Case 5} & \textbf{Case 6} & \textbf{Case 7} & \textbf{Case 8} & \textbf{Case 9} & \textbf{Case 10} & \textbf{Case 11} & \textbf{Case 12} & \textbf{Case 13} & \textbf{ACC@5} \\ \hline
Qwen2.5-7B-Instruct & 1 & 1 & 1 & 0 & 0 & 0 & 0 & 0 & 0 & 0 & 0 & 0 & 0 & 23.08\% \\ 
Qwen2.5-72B-Instruct & 0 & 1 & 1 & 0 & 1 & 1 & 0 & 0 & 0 & 0 & 0 & 0 & 0 & 30.77\% \\ 
DeepSeek-V3 & 0 & 1 & 1 & 0 & 1 & 0 & 0 & 0 & 0 & 0 & 0 & 0 & 0 & 23.08\% \\ \hline
DeepSeek-R1-Distill-Qwen-7B & 1 & 1 & 1 & 1 & 1 & 1 & 0 & 0 & 0 & 0 & 1 & 0 & 1 & 61.54\% \\ 
DeepSeek-R1-Distill-Qwen-32B & 1 & 0 & 1 & 0 & 1 & 1 & 1 & 0 & 0 & 1 & 0 & 0 & 1 & 53.85\% \\ 
QwQ-32B & 1 & 1 & 1 & 1 & 1 & 1 & 1 & 0 & 0 & 0 & 0 & 0 & 1 & 61.54\% \\ 
DeepSeek-R1 & 1 & 1 & 1 & 1 & 1 & 1 & 1 & 1 & 0 & 0 & 0 & 1 & 1 & 76.92\% \\ \hline
\end{tabular}

\label{tab:comparison}
\end{table*}

\begin{table}[H]
\centering
\caption{Diagnosis results of three LLMs under Case 4}
\scriptsize
\begin{tabularx}{0.5\textwidth}{|X|}
\hline
\textbf{Results of Qwen2.5-7B-Instruct} \\
\hline
 
\texttt{<reasoning>}
Based on the significant deviations in the sensors and the known fault characteristics, the most likely fault is **Fault Type 1 (Air Line Blockage)**, as the deviations in PT312 and PT401 align with the expected behavior of a blockage in the air line.
\texttt{</reasoning>}
\texttt{<answer>}1\texttt{</answer>} \\
\hline
\textbf{Results of DeepSeek-R1-Distill-Qwen-7B}\\
\hline
**Step-by-Step Analysis:**

1. **Identify Key Sensors**: The significant deviations in air and water flow rates, along with pressure changes, were observed in sensors like PT312, PT401, PT408, FT305, FT104, and FT406.

2. **Cross-Reference with Fault Knowledge**: The observed trends in decreasing air and water flow rates, along with pressure decreases, matched the expected behavior of a water line blockage (Fault Type 2).

3. **Determine Data Sufficiency**: The data from key sensors provided enough information to infer the fault, without needing additional data retrieval.

**Conclusion:**

The most probable fault is **Fault Type 2: water line blockage**.

\texttt{<answer>}2\texttt{</answer>}\\
\hline
\textbf{Results of Deepseek-R1}\\
\hline
\texttt{<reasoning>}
1. **Key Sensor Deviations**:\\
   - **Air Delivery Pressure (PT312)**: Decreasing trend (158-200), lower than ideal.\\
   - **Flow Rate Input Air (FT305)**: Higher than ideal (5-24, 37-200), inconsistent with a blocked air line.\\
   - **Flow Rate Input Water (FT104)**: Drops to zero (156-185) despite VC101 valve opening more (156-200 deviation).\\
   - **Pressure in Riser (PT401/PT408)**: Decreasing trends, likely due to reduced water flow affecting mixture density.\\
   - **Position of VC101**: Increasing beyond ideal values, indicating attempts to compensate for restricted water flow.

2. **Cross-Reference with Fault Knowledge**:\\
   - **Fault 1 (Air Line Blockage)**: Contradicted by FT305 (air flow rate increasing). A blocked air line would reduce FT305 readings.\\
   - **Fault 2 (Water Line Blockage)**: Matches FT104 (water flow drops to zero) and VC101 (valve opens more to compensate). Physical blockage downstream of VC101 explains these trends.\\
   - **Fault 3 (Top Separator Blockage)**: Inconsistent with FT406 (output flow increases sharply). A blocked input would reduce output flow.\\
   - **Fault 5 (Slugging)**: No oscillatory pressure/flow patterns described in observations.\\
   - **Fault 6 (2” Line Pressurization)**: Requires PT417 data, which is not analyzed here. Primary deviations align with water blockage.

3. **Conclusion**:
   The simultaneous drop in water flow (FT104) and overcompensation by VC101 (opening beyond normal) strongly indicate a water line blockage (Fault 2). The air system deviations (PT312, FT305) are secondary effects or unrelated to the primary fault.
\texttt{</reasoning>}

\texttt{<answer>}2\texttt{</answer>}\\
\hline
\end{tabularx}\label{tab:compare}
\end{table}

The results can be found in Table \ref{tab:comparison}.
The following conclusions can be drawn. First, the proposed method is effective and can achieve up to 76.92\% accuracy with only 500 normal samples without any faulty sample training. Second, the performance of the inferential macromodel is generally higher than that of the normal macromodel, which indicates that for fault diagnosis tasks, the inferential macromodel is more competent. Finally, for inference LLMs, larger model parameters result in higher accuracy.

Here we select Case 4 for analysis. We focus on analyzing qwen2.5-7B, DeepSeek-R1-Distill-Qwen-7B, and DeepSeek-R1. The former two are compared considering that they are the same model architecture and number of parameters, but the latter possesses reasoning capabilities. DeepSeek-R1 is introduced to compare with the second one to show how the number of parameters changes the performance of the reasoning LLMs.

The related results can be seen in Table \ref{tab:compare}.
For the non-reasoning Qwen2.5-7B, its output was short and got the wrong answer.
It mentioned PT312 as the reason for fault 1. If it were fault 1, the value of PT312 would be taken up, but it actually tends to go down.
For the DeepSeek-R1-Distill-Qwen-7B, it gave the correct answer and mentioned that no additional tool calls need to be made. But its reasoning process was rather sketchy.
For the Deepseek-R1, it not only got the right answer, but also gave an analysis for excluding other faults. In particular, fault 1, it mentioned that the FT305 (air flow rate) raise conflicts with the air flow blockage.

\section{Conclusion}
This study addresses the critical challenge of bridging the semantic gap between industrial time-series data and natural language understanding for explainable zero-shot fault diagnosis.
For the first time, we formally define the industrial time-series description task and propose a Signals-to-Semantics (S2S) framework to realize temporal-semantic alignment through a S2S operator and a multi-turn tree-structured diagnosis method based on LLMs.
The S2S operator successfully converts raw sensor data into concise and domain-aware natural language descriptions, including trends, periodicity, and deviations. 
Experiments are conducted on the multiphase flow process to show the feasibility and effectiveness.
The proposed framework can achieve 76.92\% diagnosis accuracy with 500 normal samples and without any fault data.
This work pioneers a new frontier in industrial AI, where temporal-semantic alignment transforms raw signals into explainable dialogues.

\vspace{12pt}
\color{red}

\end{document}